\documentclass[letterpaper]{article}
\usepackage{aaai}
\usepackage{times}
\usepackage{graphicx}
\usepackage{helvet}
\usepackage{courier}
\usepackage{amsmath}
\usepackage{algpseudocode}
\newcommand{\indep}{\perp \!\!\! \perp}

\begin{document}
%
\title{Fairness in Federated Learning for Spatial-Temporal Applications }
\author{Afra Mashhadi\textsuperscript{\rm 1}, Alex Kyllo\textsuperscript{\rm 1}, Reza M. Parizi\textsuperscript{\rm 2}\\ 
 \textsuperscript{\rm 1} Computing and Software System, University of Washington\\
 \textsuperscript{\rm 2} College of Computing and Software Engineering,  Kennesaw State University  \\
mashhadi@uw.edu, akyllo@uw.edu, rparizi1@kennesaw.edu  
}

\maketitle
\begin{abstract}
\begin{quote}
  
  Federated learning involves training statistical models over remote devices such as mobile phones while keeping data localized. Training in heterogeneous and potentially massive networks introduces opportunities for privacy-preserving data analysis and diversifying these models to become more inclusive of the population.  Federated learning can be viewed as a unique opportunity to bring fairness and parity to many existing models by enabling model training to happen on a diverse set of participants and on data that is generated regularly and dynamically.   In this paper, we discuss the current metrics and approaches that are available to measure and evaluate {\em fairness} in the context of spatial-temporal models. We propose how these metrics and approaches can be re-defined to address the challenges that are faced in the federated learning setting.
\end{quote}
\end{abstract}

\section{Introduction}


Understanding human mobility based on location-based data generated by smartphone devices has become a fundamental part of the urban and environmental planning in cities.   Through collection of these geo-traces, it has become possible    for the scientific community and policy-makers to model  citizens' daily commutes using crowd-sensed car-share data~\cite{ke2017short}, city bicycles~\cite{li2015traffic} and RFID transportation cards~\cite{silva2015predicting,mashhadi2016understanding}, or to   build  predictive algorithms to estimate people’s flows~\cite{hoang2016fccf,zhang2017deep} and community structure~\cite{ferreira2020deep,chuah2009identifying}. 

A current underlying assumption of these models is that there exists a large-scale centralized repository of location data that can be leveraged to train these models.  However, collecting large-scale location data from citizens over a long period of time is not an easy task and poses three substantial challenges. First, there are participation barriers due to privacy concerns associated with location data. Indeed, past research has shown that out of all different data types, people are least likely to be willing to share location data~\cite{gustarini2016anonymous}. Secondly, due to the enormous effort that is associated with collecting such large repositories of mobility trace, the data collection is often done infrequently. Efforts such as Mobile Data Challenges~\cite{laurila2012mobile,zheng_geolife_2010} and Data for Development challenges~\cite{de2014d4d,salah_data_2018} are an example of this type of large scale data collection that have advanced research in urban and social computing. These types of collected traces, however,   quickly become outdated and only capture a static snapshot of the citizens' behavior over a  given window of time.  Thirdly, research has shown that most centralized spatial-temporal datasets are polluted by systemic socio-economic and racial discrimination~\cite{yan2021equitensors}, and thus the algorithms that model such data need to be actively adjusted to avoid reinforcing biases, without the possibility to go collect additional data.
 
 To address the first two challenges, a handful of decentralized technologies are being experimented to address this issue, of which Federated Learning (FL)~\cite{mcmahan2016federated} is earning trust as a more promising approach. FL relies on end-device to train on their local training data and share global ML model weights with the FL central server. FL empowers the end-devices to train the local model with their local data and share the benefits of the aggregated global model from $n$ clients. In the context of smart city, the mobile crowdsensing community has recently started to explore alternatives and possibilities of a paradigm shift that would decouple the data collection and analysis from a centralized approach to a  distributed setting~\cite{jiang2020federated}

However, for such a paradigm shift to work, the research community needs to address the third challenge that is to ensure fairness is achieved in a such setting. In order to do so, a set of metrics for measuring and evaluating {\em fairness} in the context of spatial-temporal FL models are required. In this work, we examine such systems through the lens of fairness, review, and define various approaches to guarantee fairness in FL. In so doing, we rely on the methodology defined by Friedler et al.~\cite{friedler2021possibility} and examine the fairness criteria of the FL models under the two views of the world as defined based on the mechanism of mapping construct to the observed space. Specifically, we aim to answer the following research questions:  

{\em RQ1: How can the fairness metrics be defined for spatial-temporal applications in the context of privacy-preserving FL models?}

If we are able to define a set of metrics for measuring fairness based on {\em RQ1}, at which stages of the training and how can we ensure the outcome of the FL model is fair? That is:

{\em RQ2: How and what   approaches  should be applied to ensure fair FL models?}

The rest of this paper is organized as follows: we first review the related work in the context of spatial-temporal machine learning models and applications specific to the FL domain. We then cover the broader review of fairness in machine learning and describe the existing metrics and processes that are commonly used in the literature, describing the challenges of applying them to the FL setting. We then define a set of fairness criteria and metrics specific for the spatial-temporal applications and posit various strategies and stages that fairness can be embedded into FL models. 

\section{Related Work}
In this section we review the related literature categorized into two groups of related mobility works that tackle applications of smart city and mobility models, before exploring the more recent models of federated learning in the spatial-temporal applications. 
 \subsection{Smart City and Mobility Models}

Predicting dynamic urban activities such as traffic flow, visitation patterns, and public safety has become a fundamental task for the public and private sectors. For example, mobility system operators such as ride-hailing and bike-share companies often use accurate demand estimates to guide resource optimization and maximize system utility~\cite{silva2015predicting,li2015traffic}. 

The empowering component of these modern prediction systems is having access to a large amount of mobility traces of citizens paired with advanced machine learning techniques. For example, traditional time series models such as ARIMA and Gradient Boosting Regression Trees (GBRT) have been extensively used in the context of smart cities to predict urban events~\cite{hoang2016fccf,lippi2013short,ke2017short,silva2015predicting}. More recently, advanced  deep neural networks have become popular for modeling complex spatio-temporal data as mobility data has been shown to exhibit non-linearity.  For many of these models, the only assessment is measuring the accuracy of the predictions, which can help allocate city-wide resources and implement policies.

\subsection{Applications of Federated Learning in Smart Cities }

 To date, research in real-world applications of federated learning in smart cities are still in their infancy and limited to a handful of examples. Smart urban security~\cite{baig2017future} is an emerging field that has seen the most integration of federated learning in the context of smart cities.   Based on machine learning, smart security can perform post-event analysis and self-learning, constantly accumulating experience, and continuously improving pre-warning capabilities. Federated Learning offers a machine learning training scheme that allows the use of large amounts of collected data in daily applications~\cite{preuveneers2018chained}. Outside of security applications, Mashhadi et al.~\cite{IJCNN} proposed an application of federated learning for discovering urban communities. They showed that by using the GPS traces that are stored on each device and collaboratively training a deep embedded clustering model, it is possible to detect meaningful urban communities without the need for location information to be shared.

\section{Fairness in Machine Learning}

Fairness in ML is defined as either individual fairness or group based fairness. The most adopted metrics for fairness in machine learning are widely based on group based fairness which is also known as Statistical parity and Demographic parity~\cite{dwork2012fairness}. These metrics aim to ensure that there is independence between the predicted outcome of a model and sensitive attributes of age, gender, and race. Variations of statistical parity exist which concentrate on relaxation of this measure by for example ensuring that groups from sensitive attribute and non-sensitive attribute meet the same mis-classification rate (False Negative Rate) also known as Equalized Odds~\cite{hardt2016equality}, or equal true positive rate (also known as Equal opportunity~\cite{hardt2016equality}). 
On the other hand, individual fairness claims that similar individuals (with respect to a specific task) should be treated similarly with respect to that task. 
Much of the research on fairness in machine learning can be framed in an optimization context, where the goal is to maintain good predictive performance while satisfying a number of group-level or individual fairness constraints. Next to algorithmic approaches, also progress has been made with respect to theoretical analysis to better understand the possibility or impossibility of fairness with its different and often conflicting notions.  In~\cite{friedler2021possibility}, authors  proposed a framework for understanding   different definitions of fairness through two views of the world: {\em i) We are all equal ({\bf WAE}) and ii) What you see is what you get ({\bf WYSIWYG})}.  The framework shows that the fairness  definitions and their implementations correspond to different axiomatic beliefs about the world described as two worldviews that are fundamentally incompatible. 
It suggests that  if an application follows the WAE worldview, then the demographic parity metrics are relevant.  For applications that follow the WYSIWYG worldview, then the equality of odds metrics are suitable for measuring fairness.  Other group fairness metrics such as  equality of opportunity lie in-between the two worldviews and are recommended to be used appropriately.


\subsection{Holistic Fairness in Machine Learning}

 Fairness in ML can be achieved in three different stages: pre-processing the data, in-processing where the model is modified to become fair, and post-processing where the outcome of the model is tuned to meet the fairness criteria. 
 
{\em pre-processing:} Pre-processing methods concentrate on  removing representation bias and/or labeling bias in the training data in order to ensure fair balanced training data. For example, applicants of a certain gender might be up-weighted or down-weighted to retrain models and reduce disparities across different gender groups.
\\

 {\em in-processing:} %
In-processing approaches rely on adjusting the model during training to impose
fairness constraints or include fairness terms in the loss function to be
optimized. One in-processing strategy is representation learning via autoencoder
networks that are encouraged to learn features that are orthogonal to the
sensitive attribute--that is, they impose regularizing priors and/or penalty
terms to learn a latent representation $z$, which is both maximally informative
about the target variable $y$ and minimally informative about the sensitive
variable $s$. The Variational Fair Autoencoder from~\cite{louizos2017variational} and the Flexibly Fair Variational Autoencoder
(FFVAE) from~\cite{creager_flexibly_2019} are two prominent
examples of this approach. Adversarial debiasing, introduced in
\cite{zhang_mitigating_2018}, adds a discriminator sub-model to predict the
sensitive class label from $z$ and adds the negative discriminator loss to the
overall model loss function, encouraging the model to learn embeddings that are
minimally predictive of the sensitive variable.

{\em post-processing:} Finally as most of the times in-processing approaches are not possible due to lack of access to the modifying the underlying model.    Majority of the traditional linear algorithmic machine learning literature leverages   the post-processing approach. These class of fairness algorithms take an existing classifier and by treating it as a black box they aim to adjust the outcome. Such approaches also require access to the sensitive feature as input, so that they derive a transformation of the classifier's prediction to enforce the specified fairness constraints. The biggest advantage of threshold optimization is its simplicity and flexibility as it does not need to retrain the model.




\section{Achieving Fairness in Spatial-Temporal FL }

 \subsection{Fairness in Federated Learning }
 
Literature in fair federated learning is still in its infancy and often framed as equal access to effective models. That is the majority of the literature sees the goal of a Fair FL model as training a global model that incurs a uniformly good performance across all devices.   For example,  Agnostic Federated Learning (AFL)~\cite{mohri2019agnostic} and AgnosticFair~\cite{du2021fairness} optimize the worst weighted combination of local devices.  Other techniques approach fairness at the time of aggregating the models,  such as those FedMGDA~\cite{wang2021federated} and q-FFL~\cite{li2019fair}, re-weight the loss functions such that devices with poor performance will be given relatively higher weights.   These works have so far tried to define fairness as a one-fits-all and independent of the use cases and applications that the FL models are applied. For example, in the context of location data, approaches such as q-FFL that merely consider the number of data-points per device to account for fairness will fall short as in the case of location data frequency, regularity, and entropy is far closer indicators of the usefulness of the data than volume. That is two participants with an identical number of spatial-temporal data points could correspond to two distinct behavior of one being mostly stationary and the other having a very diverse trajectory portfolio (e.g., a taxi driver).
 
Furthermore, the experimental results of measuring fairness in FL literature have been arbitrary. In one study~\cite{yue2021gifair}, authors define biases in terms of the color of the handwritten digits in MNIST (blue versus black group).  Such studies fail to address biases as contextual and societal driven. We argue that biases in spatial-temporal applications are highly  driven by  manifestation of   socio-economic and  demographic geographical challenges. Thus a generic design of fairness methods  misses the required  perspective of fairness that is needed for these applications. In this paper, we aim to define a set of methods and roadmap to bring fairness to FL models that are applied in the context of spatial temporal application.  

\subsection{Fairness in spatial-temporal FL }

\begin{figure}
    \centering
    \includegraphics[scale=0.3]{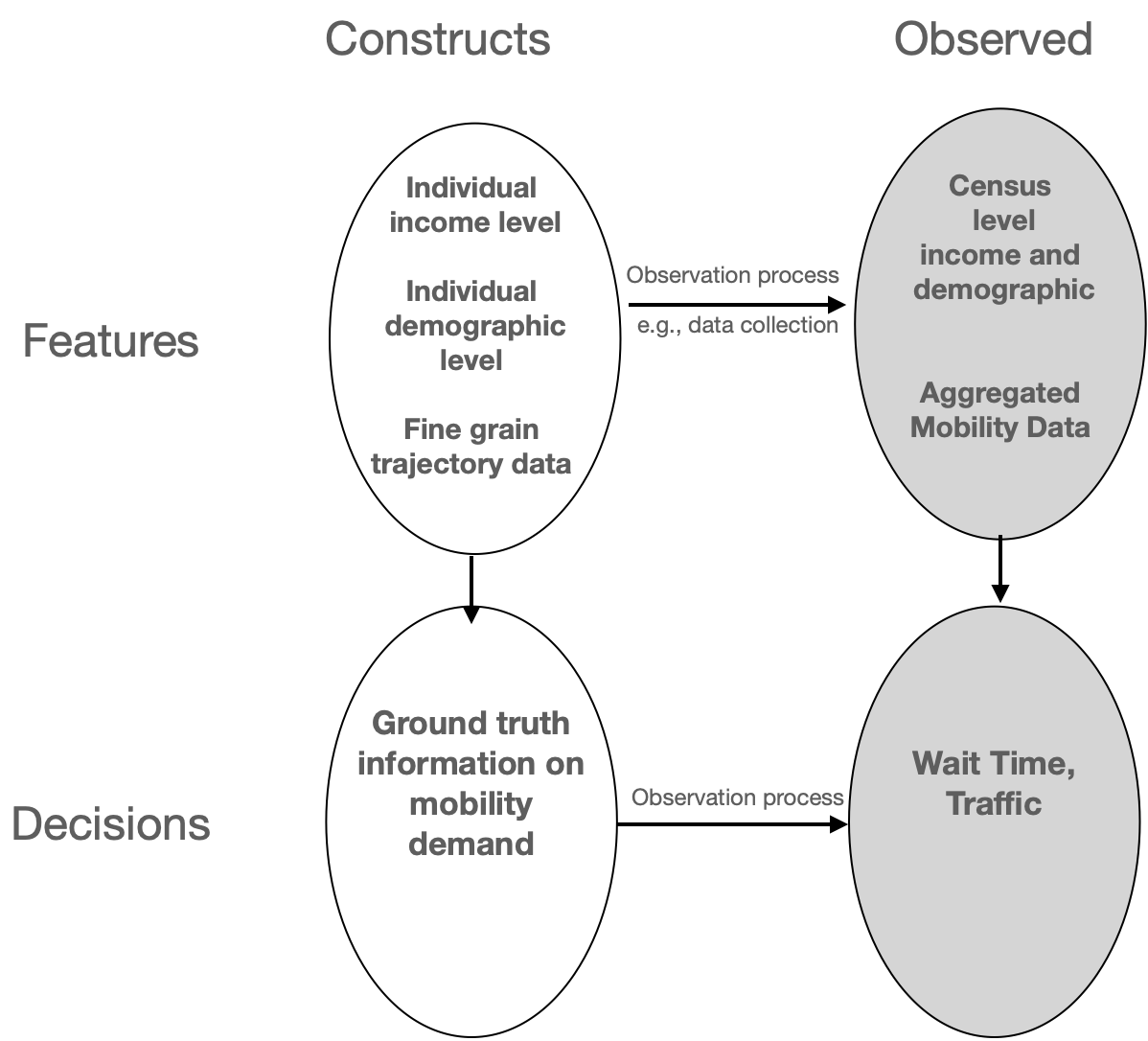}
    \caption{Design space diagram illustrating the decision making and transformations among four spaces.}
    \label{fig:desginspace}
\end{figure}

We model an algorithm making decisions about individuals as a mapping from a space of information about people, which we will call a {\em feature space}, to a space of decisions, which we will call a {\em decision space}. In the context of the spatial-temporal applications, feature space can range from demographic, socio-economic representation at the  census block (i.e., macro level)  to   citizen's fine-grain trajectory information defining  their day to day mobility across the city (i.e., micro level). The decision space is the output of the trained models which would lead to decisions in regards with resource allocation within a city.     Based on the theoretical framework offered by~\cite{friedler2021possibility}, we use the notion of $construct$ space as the idealized representation of information about people and decisions and $observed$ spaces which contain the results of an observational process that maps information about people or decisions to measurable spaces of inputs or outputs. Figure~\ref{fig:desginspace} illustrates this design space for spatial-temporal applications. More specifically it presents the following: 

{\bf The Construct Feature Space (CFS)} is the space representing the ``desired" or ``true" collection of information about people to use as input to a decision-making procedure. For example, this includes features like  individual income and demographic information and their fine-grain continuous GPS data. 

{\bf The Observed Feature Space (OFS)} is the space containing the observed information, generated by an observational process such as data collection efforts. For example, this includes the aggregated mobility information or aggregated demographic and socio-economic information as often available through Open Data repositories.

{\bf The Construct Decision Space (CDS)} is the space representing the idealized outcomes of a decision-making procedure. For example, this includes how much mobility and transportation demand is really required in different areas of the city. 

{\bf The Observed Decision Space (ODS)} is the space containing the   observed decisions from a concrete decision-making procedure. 

{\bf Challenges and Opportunities:} Within this design space, we can observe some challenges specific to the federated learning, that is the observed feature space is not visible outside of the local device and is $unknown$ to the FL server.   Such constraint  means that we need to consider new ways of applying fairness in i) pre-processing by limiting the observations to the devices, ii) in-processing by applying fair regularizers that work at global level and and ii) post-processing stages. 
In the other hand,  FL is able to reduce the gap between the observed features and the construct features as it removes the need for data collation and sharing procedure. This property allows us to   define a set of metrics that enable us to measure fairness by considering individual's trajectory. 
 Relying on these constraints and opportunities, we address the  two research questions in the subsequent sections.

\section{RQ1: Defining Fairness for Spatial-Temporal FL Models}

In this section, we offer two categories of fairness definition based on the construct mapping that we described earlier: Group fairness  ensures that groups are mapped to, on the whole, similar decisions in the observed decision space. Whereas, individual fairness  guarantees that individuals who are similar in the observed feature space receive similar decisions in the observed decision space.

\subsubsection{Individual Fairness}
The definition of individual fairness is task specific and prescribes desirable  outcomes for a task in the construct decision space. 

In the FL since group based fairness requires potentially non-observable  information regarding sensitive attribute of people,  it is most natural to define it by considering trajectory of individual people. 

In this vein, we define an FL model to {\em guarantee} individual fairness if it has same outcome (i.e., level of accuracy) for the people with $similar$ mobility behaviour. The are multiple approaches for defining similarity of mobility behavior in here we propose two measures:

{\bf Definition 1. Fano Inequality of Maximum Predictability}

 Mobility literature defines the highest potential accuracy of predictability of any individual, termed as ``maximum predictability'' ($\Pi_{max}$)~\cite{lu2013approaching}. Maximum predictability is defined by the entropy of information of a person's trajectory (frequency, sequence of location visits, etc.). 
 
 We focus on entropy and predictability analysis of day-to-day movements of individuals as recorded by their devices. Let $X_i = \{x_1, x_2, …, x_T\}$ be the sequence of daily locations for person $i$ during the data collection period of $T$ days. $x_j$ is the last observed location ID of person $i$ on day $j$. The uncertainty   of the trajectories can be then  measured by its true entropy. Larger entropy indicates greater disorder and consequently reduces the predictability of an individual's movements. We define entropy following notion in~\cite{wang2020entropy,lu2013approaching} and measure true-entropy E as:
 
\begin{equation}
E= -\Sigma_{X'_i \in X_i} P(X'_i)log[P(X'_i)]
\end{equation}
where $P(X'_i$)is the probability of finding a sub-sequence  $X'_i$ in $X_i$, considering both spatial and temporal patterns.
 
Given the entropy E for an individual i and $L_i$ distinct number of locations, Fano's inequality gives an upper limit for the predictability of individual i.

\begin{equation}
    \Pi^{max}_i=\Pi^{Fano}_i(E_i,L_i)
\end{equation}

Based on this formulation we define individual fairness in FL as:{ \em the individuals with the same level of $\Pi^{max}_i$ should receive the same outcome from FL models.} 

To achieve this notion of fairness, each device requires to calculate and share their Fano inequality value with the FL server. The FL server can then use these information to achieve fairness in multiple  ways as we will describe in the next Section.

{\bf Definition 2. Structural Similarity Index of Mobility}

Our second definition corresponds to a fine grain similarity between mobility trajectories of individuals. There are various ways of measuring distance between trajectories of people such as simple Manhattan distance, Mean Squared Error etc. In this paper, we define similarity as measured by the Structural Similarity Index Measurement ($SSIM$). In order to compute SSIM, we first transform the location  data to heatmap images by processing raw GPS traces, sequence of latitude and longitude set of coordinates over time ($T$), and creating  heatmap images that described the mobility profile of the user at that time interval.   The first step in our approach is to split the raw traces according to $T$  for each user.

In addition to the temporal variable $T$, any mobility similarity method is highly dependent on the spatial granularity of the data. That is the finer spatial granularity becomes, overall similarly among people decreases. Whereas for coarser granularity similarity will increase. 

Our  algorithm relies on two variables of  the width, $w$, and length, $l$, of the outer rectangle in miles as parameters of the study and based on a given   input cell size $C$, it calculates the pixel representation for each cell. Such that: 
\begin{equation}
    FM = 
    \begin{bmatrix}
    C_{0,0} & C_{0,1} & ... & C_{0,w} \\
    C_{1,0} & C_{1,1} & ... & C_{1,w} \\
    ... & ... & ... & ... \\
    C_{l,0} & C_{l,1} & ... & C_{l,w}
    \end{bmatrix}
\end{equation} \par

To generate images,  we normalized each value at FM(i, j) between 0 and 1 with the natural logarithm with a base of the max value present in the FM. We represent this frequency in terms of pixel intensity where the pixel intensity of $FM(i,j)$ is represented with respect to the maximum value within the FM.
 

SSIM consists of three comparison measurements to yield an overall similarity measure between two images. Luminance $\ell$ is the comparison of pixel intensity through a given point, hence the difference of ‘brightness’ from image x to y. Contrast $c$ judges an image by the distortions and texture from image x to y. Structure $s$ focuses on the patterns between pixels and carries the special information of each image, by which it compares these patterns from image x to~y.

\begin{equation}
	SSIM(x,y) = \frac{(2\mu_x \mu_y + C_1)(2\sigma_{xy} + C_2)} { (\mu^2_x + \mu^2_y + C_1 ) (\sigma^2_x + \sigma^2_y + C_2) }
 \label{eq:ssim}\end{equation}

The SSIM index is calculated on various windows, perceived qualities of each image, x and y which must share a common size N x N. Equation~\ref{eq:ssim} shows each measurement of quality where $\mu$ is the average pixel intensity of x and y respectively, $\sigma^2$ is the variance of pixel values of x and y respectively, and $\sigma$ is the covariance of x and y. The constant, $C_N$, represents a value less than 1, where $K_1$ = 0.01 and $K_2$ = 0.03, multiplied by the dynamic range of pixel values contained within an 8-bit grayscale image where ${L = 2^{255} - 1}$. The final expression of $C_n$ is denoted by the expression ${C_n = (K_n L)^2}$. This value is included to avoid division by 0 in the case where ${\mu_x^2 + \mu_y^2}$ are very close to zero. Based on this formulation we posit an individual  fairness definition in FL as: {\em the individuals with same SSIM level receive the same outcome from the globally aggregated model.}

\subsubsection{Group Fairness}
In the context of spatial-temporal applications group-based fairness definitions are rare to find.~\cite{yan2020fairness} et al. defines fairness in terms of,  region-based fairness gap which assess  the gap between mean per capita ride sharing demand across groups over a period of time. The two metrics differ from each other with one being based on binary label associated with the majority of the sub-population (e.g., white) versus a continues distribution of the demographic attributes. To the best of our knowledge~\cite{yan2020fairness} is the only work in literature which offers a  group-based fairness metric for spatial-temporal data.  However, their work focuses on per area demand prediction, where as in the context of our study we are focused on data generated by individuals devices and reflects historical mobility  information of the user. Assuming the prevalence of such data, we can define group-based fairness as conditional independence:
 
\textbf{Definition 1.}{\em The FL model, F(x), satisfies the group fairness if $C \indep  A | Z $. }
 where C is the outcome of the model F, A is the sensitive attribute and Z is the  mobility characteristics of the user or community. In the case of binary classification, where the outcome of the $F$ is the global model works for certain individuals denoted as C=1 (i.e., acceptance), conditional independence is formulated as: 
 
 \begin{equation}
 \frac{P\{C = 1 | A = a, Z=z\}}{P\{C = 1 | A = b, Z=z\}}= 1 - \epsilon.
 \end{equation}
 
 That is for all groups a,b with similar z mobility characteristics/resources, the global federated model must work almost equally with a positive amount of slack $\epsilon$\footnote{Previous work argues that when $\epsilon=0.2$ this condition relates to the 80 percent rule in disparate impact law~\cite{feldman2015certifying}}.   To achieve the defined group-based fairness we need to define what are the sensitive attributes and the mobility condition. We can define both these categories in macro and micro scale. In this context,  macro corresponds to the characteristics as observed by a sub-population whereas micro corresponds to the individual users.

 \begin{description}
 \item[Sensitive Attribute:] At macro level various sensitive attributes can be defined that correspond to the characteristic of a community. Such groups could for example be formed based on socio-economic vulnerability index as published by CDC in census block group (CBG) granularity, or based on demographic parity by looking at census-tract level minority status information. In the context of FL, the home location of the participants devices could be inferred and mapped to census block group data, to assign a sensitive attribute to each user. 
 
 At micro level, the sensitive attributes such as income level and demographic information could be collected directly from the participant. 
 
 \item[Mobility Condition:] At macro level, the mobility characteristics could for example correspond to the availability and dependency on the  transportation at each census block group. Data such as the {\em mobility and transportation vulnerability index}, as defined by CDC, can serve as this condition. At micro level, such information can correspond to the number of data records.

 \end{description}
 
 \begin{figure*}
    \centering
    \includegraphics[scale=0.3]{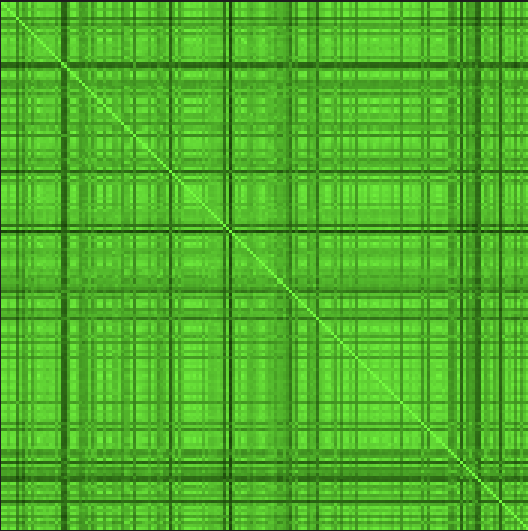}
        \includegraphics[scale=0.3]{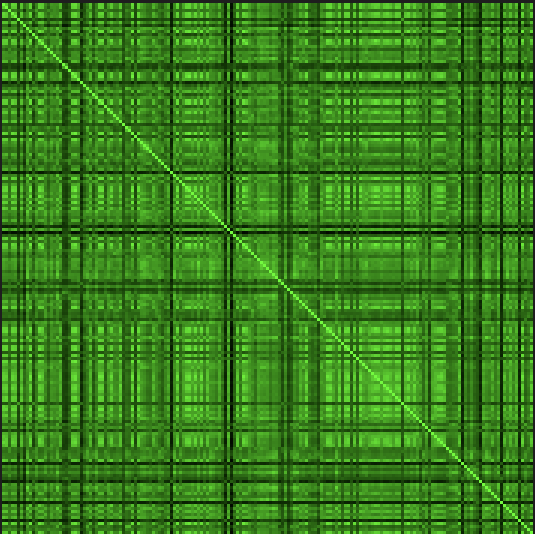}
    \includegraphics[scale=0.3]{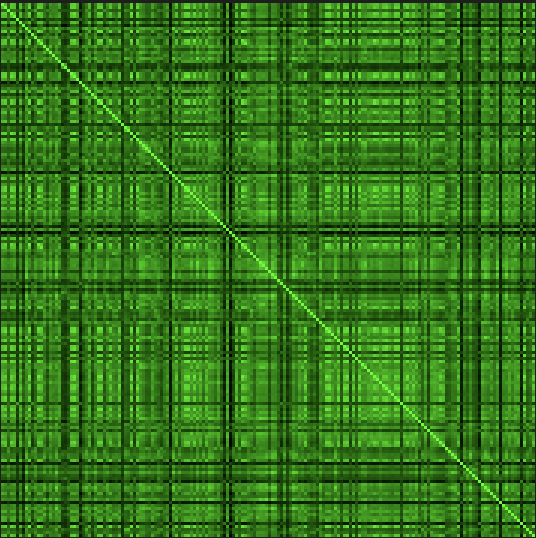}
    \caption{Structural similarity of users in MDC dataset~\cite{laurila2012mobile} for various spatial granularity of 500 meters down to 100 meters. The brighter color presents higher similarity. As the granularity becomes finer, the mobility similarity between the users decreases. 
    }
    \label{fig:MDC}
\end{figure*}

Putting these metrics together, group-based fairness enables us to argue that users/communities with the similar mobility trend should receive similar outcome of contributing to FL models, regardless of their socio-economic or demographic, corresponding to $WAE$ view of the world. The individual fairness allows to reason that  individual users with similar mobility patterns should receive the similar outcome of the FL model capturing the  $WYSIWYG$   view of the world.

\section{RQ2: Stages of Applying Fairness in FL}
In this section, we discuss possible ways of applying fairness in to models by considering the requirements and challenges of the federated learning systems.

\subsection{Pre-processing} Applying fairness in pre-processing stage often includes data augmentation and debiasing of the input to account for a more representative distribution of the population. In the FL setting, given that the data stays local on the devices and is not visible to the FL server, any pre-processing intervention would translate into participant selection strategies. That is for any FL  model to be fair and inclusive, it must be trained on a diverse representation of the data, requiring a diverse set of participants to be included in the training stages. Thus, a participant selection strategy that selects devices based on their group membership to achieve either group-based or individual fairness is required. 

To achieve group-based fairness in this stage, the FL server can indicate the number of required participants from each group and select devices uniformly across the groups. Existing algorithms such as those that aim to select participants based on incentives or resources (e.g., battery, WiFi, etc) can be used to meet these selection criteria.  

To achieve individual fairness in this stage, additional computation at the device level is required. That is first the notion of similarity amongst individuals needs to be calculated and groups of similar individuals to be detected.
 As this notion of fairness requires access to devices location data in order to compute the one-to-one similarity across users, it requires any FL model to be embedded with a federated clustering algorithm to first discover the clusters of similar individuals and then apply participant selection  for ensuring fairness amongst members of same cluster.

 Approaches such as Deep Federated Clustering~\cite{IJCNN} for unsupervised learning and IFCA~\cite{ghosh2020efficient} for supervised learning (when demographic labels are present at device level) could be applied to discover clusters of similar users. 
 
 To validate this participant selection strategy, we used Mobile Data Challenge(MDC) Dataset~\cite{laurila2012mobile}. The MDC dataset contains a large variety of data collection sensors including an accelerometer, phone records, wireless access points, Bluetooth connection, and GPS recordings. For the purpose of this work, we are only concerned with the geo-location data as time-series records. The mobility traces were contributed by 185 users from 2009 to 2011 sampled 6 times per second. We applied the FM algorithm described earlier to create heatmaps based on each user's trajectory.   Figure~\ref{fig:MDC} illustrates the SSIM matrix for users in this dataset for various spatial granularity. As the spatial granularity becomes coarser, the users become more similar to each other, impacting the number of distinct groups that can be discovered.  

\begin{figure}
    \centering
    \includegraphics[scale=0.3]{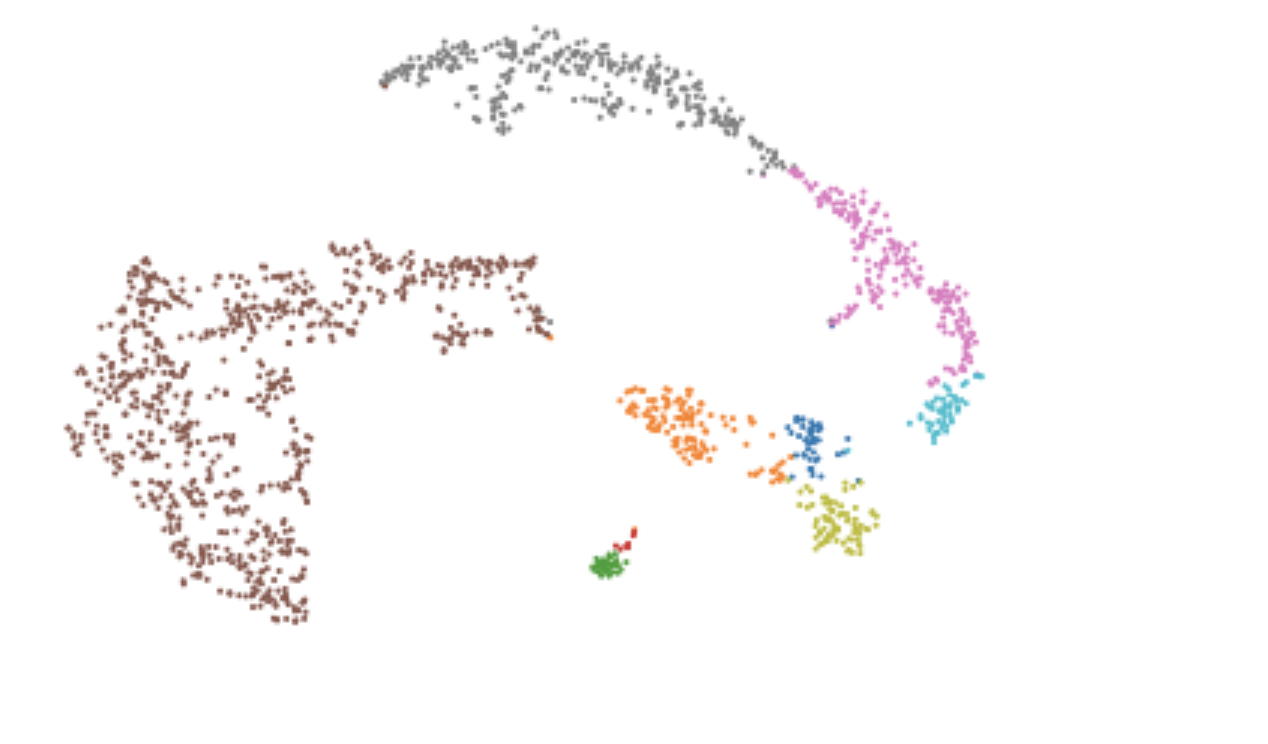}
    \caption{Discovered similar clusters of users for MDC dataset where each users in each cluster have a high SSIM similarity between them and low SSIM similarity between the clusters. }
    \label{fig:clusters}
\end{figure}

Figure~\ref{fig:clusters} presents the   clusters of MDC users  that are discovered to be similar to each other based on the federated DEC algorithm as presented in~\cite{IJCNN} for the spatial granularity of 500 meters after 10 epochs of local training. This algorithm relies on representation learning to optimize a convolutional autoencoder for two optimization tasks of learning the representation of the images from each user (minimizing reconstruction loss) and maximising the clustering by adding a clustering layer that aims to reduce KL-divergence.   The results in Figure~\ref{fig:clusters} shows the viability of selecting the participants based on their cluster membership where users with similar SSIM should be treated similarly.  Moreover, the federated DEC  algorithm has been shown to converge in approximately 2 minutes (10 epochs of training) and consume less than 5\% memory on ordinary smartphones (with negligible energy consumption)~\cite{IJCNN}.   

\subsection{In-processing}

In-processing approaches rely on adjusting the model during the training to enforce fairness goals to be met and optimized for in the same manner as accuracy is.  This is often achieved through adversarial networks or fair representation learning approaches such as~\cite{hu2020fairnn}, model induction, model selection, and regularization~\cite{yan2020fairness}. Such approaches in the traditional ML however would not be applicable to federated learning as they rely on centralized datasets that would collectively include  a wide  representation of the data from users from different groups. In the context of FL each round of local training is performed on local data that belongs to individual users and thus it is not possible to apply regularization techniques without further intervention. One possible approach to leverage the existing in-process  approaches in FL is to rely on the FL server to send additional data to local devices at each round of training, where this data can be merged with local data to enable regularization to happen. However, such approach will be very cost inefficient and will incur a large amount of communication across the server and the devices. Alternatively approaches   could rely entirely on the local data but achieve in-process fairness through   parameter-tuning by searching the most optimal and fair parameters for each device. We believe such approaches are  highly desired for the FL community and are vital to shape the   future roadmap of    Fairness in  FL.  



\subsection{Post-Processing}

In the context of FL, we refer to post-processing as the set of processes that occur on the FL server after each round of local training.  Thus, accounting for fairness in aggregating the weights of the trained models could serve as a natural choice. For example, different aggregation methods at the server where different weights based on either group membership or similarity can be taken into account when aggregating the models~\cite{huang2020fairness}. Alternative to aggregation approaches, post-processing stage could be also considered as an opportunity to audit the trained models using synthetic data. That is at the end of each round of training on local devices, the FL server could evaluate the accuracy and fairness of the model on a centralized dataset that includes the sensitive attributes.  GAN based approaches  can be used  to create fair representation of the previously collected spatial dataset for this auditing purposes. 
Such auditing can serve as a feed-back loop into informing the distribution of the participants for the next round of training (pre-processing) and the hyper-parameter setting of the model (in-processing).

\section{Conclusion}\label{sec:conclusion}
In this paper, we defined a set of metrics for measuring the fairness of the spatial-temporal models based on the observed and construct feature space. We showed how FL can help by removing some privacy concerns and allowing for more detailed observable features space. Relying on fairness literature we defined fairness in macro and micro levels for mobility data and showed how FL models can account for fairness in three stages. Our study calls for the research community to pursue future research in creating novel approaches in applying in-process fairness into FL models. We expect to see more application specific works and studies that will leverage pre and post-processing approaches defined here to gaurantee fairness. Finally, we believe conceptual works such as ours are particularly needed to enable discussion and conversation around how to measure fairness in a different context and how to create models that  are audit-able   when trained in the FL setting. We believe topic of fairness and transparency in FL is to be a vital topic for AAAI community to pursue and explore.

\bibliographystyle{aaai}
\bibliography{ref}

\end{document}